# Automated Image Color Mapping for a Historic Photographic Collection

Taylor Arnold[1,†], Lauren Tilton[2,*,†]

[1]*Data Science & Linguistics, University of Richmond, U.S.A*
[2]*Rhetoric & Communication Studies, University of Richmond, U.S.A*

#### Abstract
In the 1970s, the United States Environmental Protection Agency sponsored *Documerica*, a large-scale photography initiative to document environmental subjects nation-wide. While over 15,000 digitized public-domain photographs from the collection are available online, most of the images were scanned from damaged copies of the original prints. We present and evaluate a modified histogram matching technique based on the underlying chemistry of the prints for correcting the damaged images by using training data collected from a small set of undamaged prints. The entire set of color-adjusted Documerica images is made available in an open repository.

## 1. Introduction

Many of the most important environmental laws and federal agencies in the United States came into existence during the large-scale political and social environmental movement that formed during the 1960s and 1970s [8]. Significant political advances that continue into the present day include the Clean Air Act (1963), the Clean Water Act (1972), and the Resource Conservation and Recovery Act (1976). The United States Environmental Protection Agency (EPA) was founded in 1970, to manage, advocate, and set standards for these new legal frameworks [21]. The EPA continues today with a staff of over 16,000 and a budget of over $12 million [3].

Documerica was an EPA-funded project running from 1972 to 1977 that aimed to "photographically document subjects of environmental concern in America during the 1970s" [1]. The photographs captured a wide range of topics. Mixed through images of water pollution, chemical spills, and factory smoke plumes are pastoral landscapes from national parks around the country [7]. Urban images of junkyards and trash are juxtaposed with images of cleaner technologies, mass transit, and Americans of all ages at play [22]. Over 15,000 photographic prints from the collection have been digitized and made available through the National Archives of the United States. As a product of the federal government, these images are in the public domain and serve as a rich potential source of documentary evidence of the United States in the 1970s and the early years of the modern environmental movement.

Unfortunately, a damaged set of Documerica prints was used for digitization. The scanned photographic prints have an intense red shift, causing everything from the sky, to lakes, to trees to have a red/orange hue in place of their normal expected colors. The sharpness of the images has not been significantly affected indicating that the damage was due to a slow degradation over time through a combination of heat and light. The color shift is sufficiently strong to reduce the aesthetic and rhetorical power of the images that are available online [12]. Interestingly, the National Archives holds two additional sets of Documerica prints. These two other sets reveal the aesthetic qualities of the original photographs and open the door to the possibility of correcting the damaged colors in digital prints.

In this article, we present and evaluate an algorithm to automatically correct the color of the damaged digitized Documerica images using a small set of the undamaged prints as training data. Our goal is to restore the aesthetic qualities of the images rather than the impossible task of perfectly matching the colors in the reference images [4, 17]. Through several quantitative and qualitative analyses, we show that our transformed images more closely represent the expected colors of the photographed scenes. Our technique is sufficiently tractable that it could be applied to other color-shifted collections that do not have a reference set to train against.

## 2. Data

The photographers employed by the Documerica project took photographs using the Kodachrome color reversal film produced by Eastman Kodak [7]. Unlike many other technologies for color photography, Kodachrome film used a subtractive technology that recorded light by measuring the amount of cyan, magenta, and yellow dye needed to print a reconstruction of the image [19]. A special setup was required to create prints from the images. Special labs were able to take film and turn it into color photographic slides through three differently colored developers [6].

The National Archives of the United States has three copies of prints from Documerica: (1) archival copies held in cold storage and available on special request, (2) a non-circulating set held in their print room, and (3) a damaged set of prints that were used for the digitization process.[1] The authors visited the National Archives and selected a set of 23 slides from across the collection to manually re-scan from both the cold-storage and non-circulating print set. We selected images that contained a variety of different colors. After determining that the cold storage images were indistinguishable from the print room copies, we decided to only re-scan the latter. We then manually cropped each of these images to match the digitized prints. One selected image had already been replaced with a corrected scan online. This left a training set of 22 images for which we had manual reference scans from the print room and the damaged,

---

[1] It is not entirely certain how the latter set was damaged or why it was selected for digitization. The third set was previously used as a circulation copy, and was likely damaged due to its continued circulation. Personal communications with the current archivists suggest that because the digitization was done off-site, someone selected the circulating copy as to send off before realizing how much it had become damaged.

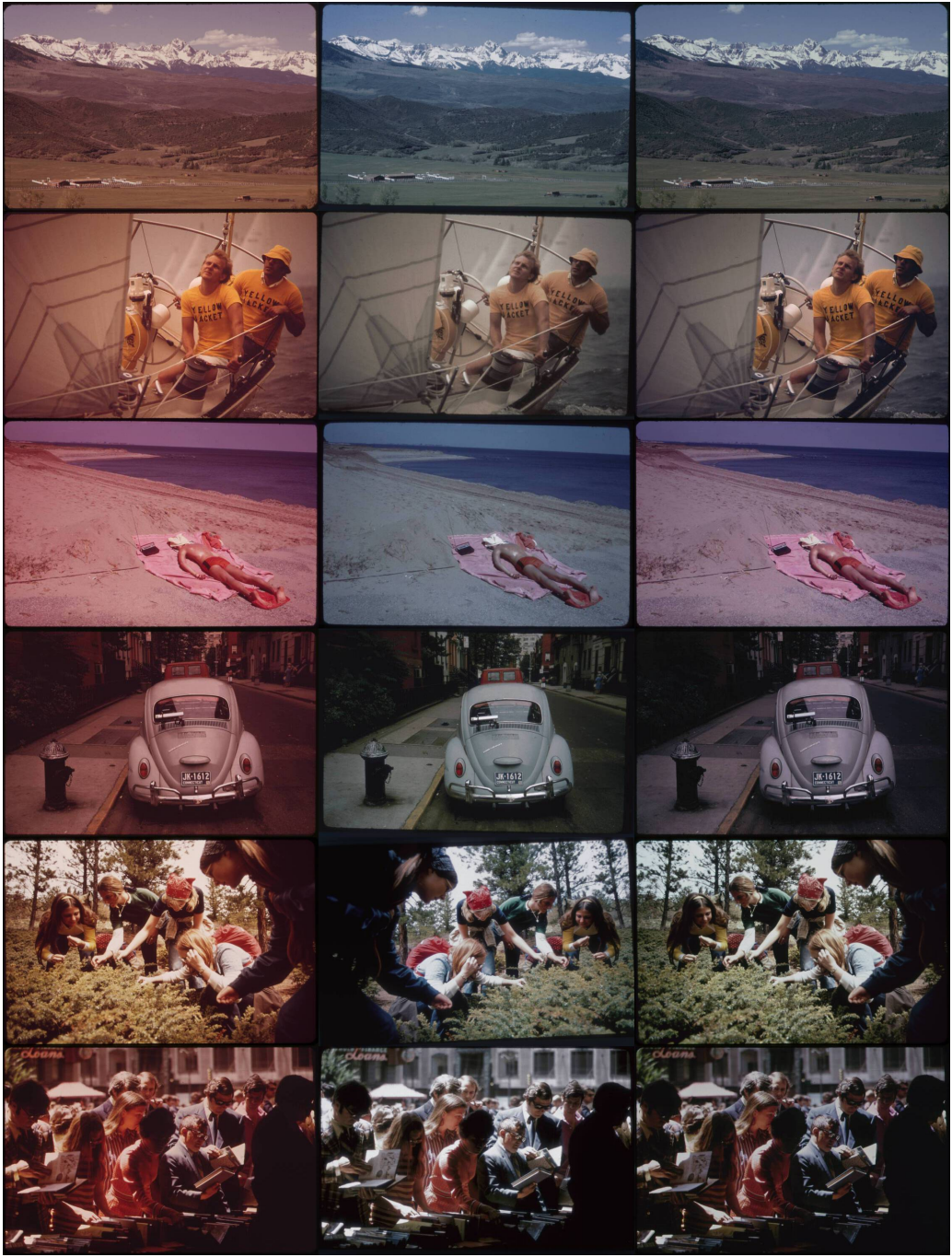

**Figure 1:** Set of selected photographs from the Documerica collection. The first column contains digitized damaged prints available online; the second column shows manually scanned from the undamaged prints available at the U.S. National Archives; the final column shows the automatically corrected images using our method.

digitized copies on National Archives website.

## 3. Overview of Color Mapping Algorithms

Color adjustment of digital photographs is a well-studied topic and an essential element of the workflow of modern photographers. For example, a common technique for professional photographers is first to photograph a color reference card containing a variety of boxes with known hues. Then, commercial software can use this reference image to learn how to adjust any other images taken with the same equipment in similar lighting conditions [23, 24]. Adobe Photoshop provides automated algorithms for matching color across images, which is useful for applications such as putting together images from a photoshoot done under various lighting conditions [9]. Color adjustment is also an essential stylistic element for photographers [16]. Commercial photo editing software such as Adobe Lightroom provides a variety of methods for automatically applying a suite of filters to a set of images—a technique commonly used, for example, by wedding photographers to create a distinct style—or for manually adjusting the color of an individual image.

In addition to commercial software for color adjustment, there has also been considerable academic research on the specific task of matching the color between two images, a process known as *color mapping* or *image color transfer*. In one of the earliest studies of color mapping, Reinhard et al. introduced a technique consisting of a simple standardization of color intensities in a specific color space [20]. They showed that this technique worked well as both a mild correction to standardize images of the sun setting over the horizon and more drastic applications of style transfer from an oil painting to a modern photograph. In a more recent survey, Faridul et al. provide a nomenclature of available techniques for color transfer: geometry-based, statistical, and user-assisted [10]. Statistical techniques extend the ideas of standardizing the mean and variance of color intensities to more involved transformations known as *histogram matching* [18]. Much of the sophistication of novel methods has focused on making differential transformations of various types over various parts of the image [13]. One motivation for localized changes is adjusting lighting conditions as a pre-processing step for other algorithms.

## 4. CMY Median Histogram Matching (MHM)

Our approach to image color mapping with the Documerica collection is closely related to the well-known histogram matching technique. In this technique we attempt to match the distribution of colors in one image with the distribution of colors in a reference image through a monotonic transformation of the pixel intensities [14, 18]. Two special considerations in our application require some minor changes to the standard histogram matching algorithm.

First of all, through our understanding of the materiality of the prints, it is likely that the color shift that damaged the prints likely affected the images along the dimensions of the three developer colors. The contributing elements of heat, light, and/or humidity would affect each of these dyes differently according to their chemical composition. Our qualitative analysis of the consistent red shift in the digitized images further indicates the need for a color correction applied individually to each of the CMY color channels. So, in contrast to the standard strategy for histogram matching that uses color spaces adapted to the sensitivity of the human eye, we will apply our directly to the RGB/CMY pixel intensities [25].[2]

The second difference from a standard histogram matching algorithm is that we want to avoid overfitting to the distribution of a single image. In fact, one of the goals of our transformation is to restore the diversity of lighting and colors that are in the source materials. Instead of perfectly matching the distribution for a single image, we want to create a transformation by averaging the histogram transformations across the entire training set of 22 pairs of images.

The goal of our adapted technique, which we will refer to as *median histogram matching* (MHM), is to learn three monotonic functions $\hat{f}_C$, $\hat{f}_M$, and $\hat{f}_Y$ that each map an input color intensity (from 0 to 1) into an output color intensity on the same scale. Applying these to the cyan, magenta, and yellow components of a damaged image should result in a transformed image that more closely represents the undamaged form of the print. In addition to being motivated by the chemistry of the prints, the transformation mirrors tools available in popular photo editing software such as the Photos application of macOS and Adobe Photoshop.

Differences in the technology available for digitizing the undamaged prints—including the lighting, orientation, overscan size, and print-specific artifacts—made it infeasible to line up our reference images pixel-by-pixel. Following related work on color correction, we focused on lining up the general distribution of each color channel intensity. For each image $i$ and color channel $j$, we computed the percentiles for the color intensities of both the damaged image and the scanned undamaged print. Then, we constructed a function $\hat{f}_j^i$ by matching the two sets of percentiles to one another. For example, if the 20th percentile of the damaged image had a cyan intensity of 0.6 and the undamaged percentile was 0.82, $\hat{f}_j^i(0.6)$ would be set to 0.82. We then set $\hat{f}_j^i(0)$ to 0 and $\hat{f}_j^i(1)$ to 1 and filled in the intermediate values through linear interpolation. Then, the final predictions for each color channel $j$ ($\hat{f}_j$) are given by taking the median across all of the images in our training collection. These transformations are guaranteed to be monotonic and to be well-defined for all input values.

## 5. Results

### 5.1. Transformations

Figure 2 shows the learned MTM transformations derived from our training set and the method outlined in the previous section. The grey lines show the estimates $\hat{f}^i$ for each image $i$, with the darker colored lines showing the final learned transformations based on the median across the entire training set. These transformations were also applied in the final column of Figure 1.

The cyan color channel transformation is the most different from an identity mapping. The transformation suggests that the amount of cyan intensity should be increased, which aligns with our qualitative analysis of the collection has having a general red tint (the intensity of cyan can be computed by taking the inverse of the red pixel intensity). The transformation indicates that the cyan dye as degraded at a rate faster than the other two dyes. The magenta and yellow curves, on the other hand, are much closer to the identity transformation. While the algorithm does indicate some changes for individual images, these are relatively minor when we take the median across the entire training set.

---

[2]The representation of a pixel in CMY space can be computed by simply subtracting the RGB representation from 1. So, matching the CMY distribution is equivalent to matching the RGB representation.

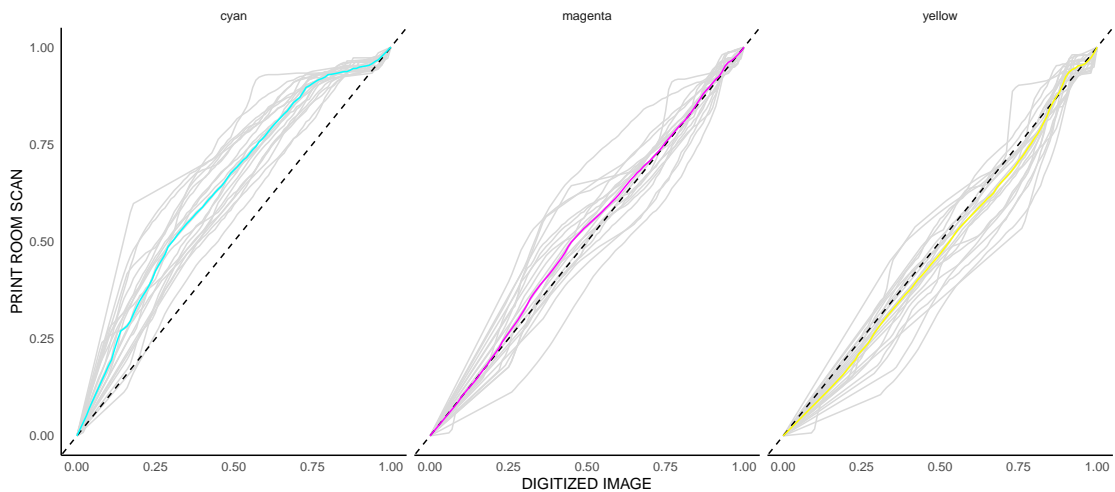

**Figure 2:** Learned MHM transformations of the CMY color channels. Grey lines show the estimates of individual images; the colored lines are the medians across the entire training set.

We see that there are considerable differences across training images. These differences are less pronounced than they first appear because over half of all the pixel intensities in the digitized images have intensities over 0.8. Looking at the upper part of each curve, we see that these are much more stable than the rest of the transformation. Additionally, the transformations on the bottom of the curves correspond to parts of the image that have a small amount of the given color dye. Often these are very bright parts of the image that are close to white, and the differences indicated on the chart have a small visual effect, as we see in the following subsection.

The differences across training images highlight two points of caution. First, the adjustments are nowhere near perfect. The final exact colors can still be quite different than the original prints, the negatives themselves, or the light reflected by the objects being photographed. Secondly, in order to find a good transformation for another collection. Taking the median across several hand-selected transformations will likely work well, whereas applying a filter trained on only one image to a larger collection can quickly lead to overfitting.

### 5.2. Leave-One-Out Cross-Validation

We can use quantitative measurements to see how well the learned median transformations work on the training data itself. In order to do this, we use a leave-one-out cross-validation technique in which we compute the median transformations with the $i$th image removed—which we denote by $\hat{f}_j^{-i}(x)$—and then compare this to the transformation given by using only the $i$th image itself [11]. To compare these two functions, we use a distance metric $d$ given by

**Table 1**
Average errors (sum of the $L^2$ norms across color channels, multiplied by 100) for predicting the color channel transformation on the training set of images. Standard errors are given after the means. Uniform errors are computed equally across the color intensities. Weighted errors use the empirical distribution for each image.

|  | Uniform Error | Weighted Error |
| --- | --- | --- |
| Identity Transformation | 3.375 ±0.400 | 3.021 ±0.301 |
| Leave-One-Out Estimator | 1.154 ±0.227 | 0.990 ±0.217 |

**Table 2**
Average distance between a set of manually transformed images by a Wikipedia editor and the transformed images through our median estimator. Distances are the average Euclidean distances per pixel. The UV and AB color spaces use the same coordinates as the CIELUV and CIELAB color spaces, but compute distances without the perceptual lightness component.

| Color Space | CIELUV | UV | CIELAB | AB |
| --- | --- | --- | --- | --- |
| Identity Transformation | 25.78 ±2.04 | 23.62 ±2.08 | 1239 ±115 | 1097 ±126 |
| Median Estimator | 16.77 ±1.20 | 12.95 ±1.59 | 602 ±49 | 471 ±56 |

the sum of the $L^2$ norms from each of the three color channels:

$$d\left(\hat{f}^i; \hat{f}^{-i}\right) = \sum_{j \in \{c,m,y\}} \int_0^1 \left|\hat{f}_j^i(x) - \hat{f}_j^{-i}(x)\right|^2 dx$$

To get have a sense of the scale of this metric, we will compare it with using the identity transformation to estimate $\hat{f}^i$. This metric treats all input color intensities equally. We also compute a weighted version of the metric weighted by the density of the colors in the image itself to provide a more accurate measurement of how much the two transformations differ on a specific image.

The results of the leave-one-out cross-validation are shown in Table 1. The cross-validated median estimator has an error rate about three times smaller than the error from using the input images. The errors are slightly smaller when weighted by the empirical distribution of the intensities, but the relative relationships remain similar. Only 1 of the 22 training images has a cross-validation error larger for the median estimator than the identity transformation, with every other image being improved by a factor of at least 50%.

### 5.3. Comparison to Manual Edits

One of the motivations for the work in this paper was our perceived ability to make the digitized, damaged Documerica prints look much closer to what we expected through manual color adjustments in commercial photo editing tools. One editor on the English version of Wikipedia, posting under the username Hohum, had a similar experience with the Documerica photographs. In January 2023, they uploaded manually corrected versions of 14 Documerica photographs to Wikipedia. No description of the method used was given. It seems unlikely that the user had access to the undamaged prints; they have uploaded thousands of other corrected images under a large variety of categories on Wikimedia Commons. While we should not treat the edits by Hohum as a gold-standard transformation that we should aim to replicate perfectly, comparing our method to the manually edited photographs offers further quantitative evidence for the efficacy of our automated technique.

As we did in comparing the transformations in the leave-one-out cross-validation, we will look at the average distances between the reference image (here, the corrected version uploaded to Wikimedia Commons) to both the uncorrected image and our automatically corrected image. Here, though, we look at the average Euclidean distance at the level of individual pixels. Because we are interested in measuring the visual perception of the transformation, we will first convert the images into two different color spaces designed to represent human visual perceptions: CIELAB and the CIELUV [15]. We report the results with and without considering the lightness dimension of the colors.

Table 2 compares our transformation to the independent, manual transformations performed by Wikipedia user Hohum. Across all four selected color spaces, the median transformation method is roughly twice as close to the manual adjustment than the original image. For all the photos, the distance from the median transformation is closer to the manual transformation in the CIELAB and AB color spaces. Only one image, which consists almost entirely of a field of monochromatic corn shot from above, has a worse CIELUV distance in the median transformation.

### 5.4. Qualitative comparison

By looking at a subset of the transformations applied outside of our training set, we can qualitatively describe how well the algorithm produces more realistic colors in the images. Creating more realistic images that are more aesthetically interesting and engaging is the ultimate goal of our work. Figure 3 shows 18 Documerica images and result of applying our transformation. The red shift in the original images is very noticeable in the first and third columns. The transformed images lose this red tone and become more color-appropriate. The image of the two people fishing and the group of people on a bus both show no elements that still have any red hues. On the other hand, images such as the sunset and the rocky cliff with a bridge still retain their expected red characteristics. In other words, we have not overcorrected the images by making them all too blue. Instead, the updated collection shows a diverse set of colors that help to re-establish the true scale and scope of the Documerica project itself.

## 6. Conclusion and Future Work

The Documerica project produced a large set of historically important documentary photographs [5]. With the increased present-day attention to environmental issues, the collection has the potential to have a visible role in helping to understand the longer history of environmentalism in the United States. However, the unrealistic red shift of the digitized collection has until now reduced its aesthetic and rhetorical appeal and thus its general reach. Through the adjustment of this shift through the algorithmic color transfer outlined in this paper and the publication of the corrected images, we hope to help rectify this situation.[3] As a next step—or, more accurately, the original motivation for this work—we plan to build an interactive digital public interface to help explore and understand the Documerica collection across its various spatial, temporal, and visual components [2].

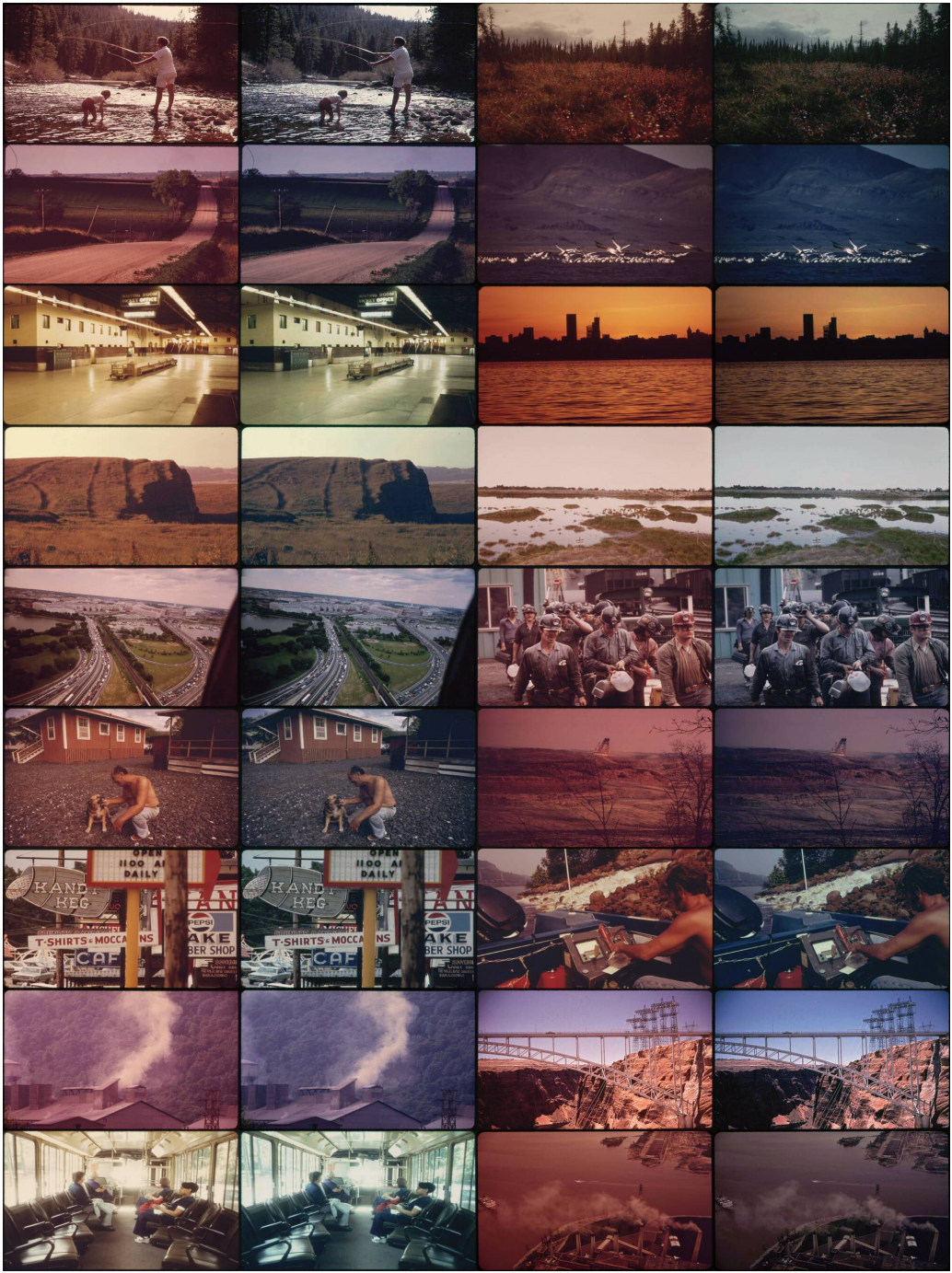

**Figure 3:** Set of selected photographs from the Documerica collection, with the digitized damaged prints available online to the left of the automatically corrected images using our method. Images are scaled to have the same aspect ratio to make them easier to display.

---

[3] To download the color-adjusted collection, see: https://distantviewing.org/downloads.